\documentclass[letterpaper, 10 pt, journal, twoside]{IEEEtran}
\usepackage{amsmath,amsfonts}
\usepackage{array}
\usepackage{textcomp}
\usepackage{stfloats}
\usepackage{url}
\usepackage{verbatim}
\usepackage{graphicx}
\usepackage{cite}
\usepackage{times}
\usepackage{epsfig}
\usepackage{amssymb}
\usepackage{amsmath}
\usepackage{booktabs}
\usepackage{comment}
\usepackage{multicol}
\usepackage{multirow}
\usepackage{caption}
\usepackage{subcaption}
\usepackage{animate}
\usepackage{tikz}
\newcommand{\norm}[1]{\left\lVert#1\right\rVert}
\makeatother
\def\eg{\emph{e.g}.} 
\def\ie{\emph{i.e}.}

\makeatletter

\usepackage[pagebackref=true,breaklinks=true,letterpaper=true,colorlinks,bookmarks=false]{hyperref}
\usepackage[capitalize]{cleveref}

\begin{document}
%
\title{Toward Reliable Human Pose Forecasting with Uncertainty}
%
%
%

\author{Saeed Saadatnejad, Mehrshad Mirmohammadi$^{*}$, Matin Daghyani$^{*}$, Parham Saremi$^{*}$, Yashar Zoroofchi Benisi$^{*}$, Amirhossein Alimohammadi$^{*}$, Zahra Tehraninasab$^{*}$, Taylor Mordan and Alexandre Alahi%
\thanks{The research was conducted in VITA laboratory at EPFL. For inquires, please contact: \tt\footnotesize {firstname.lastname}@epfl.ch}%
\thanks{$^{*}$ Equal contribution as the second authors.}%
}%
%
%

\markboth{IEEE Robotics and Automation Letters. Preprint Version.}
{Saadatnejad \MakeLowercase{\textit{et al.}}: Human Pose Forecasting with
Uncertainty} 

%


\maketitle

\begin{abstract}
Recently, there has been an arms race of pose forecasting methods aimed at solving the spatio-temporal task of predicting a sequence of future 3D poses of a person given a sequence of past observed ones. However, the lack of unified benchmarks and limited uncertainty analysis have hindered progress in the field. To address this, we first develop an open-source library for human pose forecasting, including multiple models, supporting several datasets, and employing standardized evaluation metrics, with the aim of promoting research and moving toward a unified and consistent evaluation.
Second, we devise two types of uncertainty in the problem to increase performance and convey better trust:
1) we propose a method for modeling aleatoric uncertainty by using uncertainty priors to inject knowledge about the pattern of uncertainty.
This focuses the capacity of the model in the direction of more meaningful supervision while reducing the number of learned parameters and improving stability;
2) we introduce a novel approach for quantifying the epistemic uncertainty of any model through clustering and measuring the entropy of its assignments.
Our experiments demonstrate up to $25\%$ improvements in forecasting at short horizons, with no loss on longer horizons on Human3.6M, AMSS, and 3DPW datasets, and better performance in uncertainty estimation. The code is available \href{https://github.com/vita-epfl/UnPOSed}{online}.
\end{abstract}



\section{Introduction}
\label{sec:intro}

\IEEEPARstart{H}{uman} pose forecasting consists in predicting a sequence of future 3D poses of a person, given a sequence of past observed ones. 
It has attracted significant attention in recent years due to the critical applications in autonomous driving~\cite{saadatnejad2024socialtransmotion}, human-robot collaboration~\cite{duarte2018action, vianello2021human}, robot navigation~\cite{chen2019crowd}, and healthcare~\cite{wagner_targeted_2018-1}.
The field is now witnessing an arms race of forecasting models using different architectures that have shown increasing performances~\cite{mao2020history, sofianos2021stsgcn, ma2022progressively}.

\begin{figure}[!t]
    \centering
    \scalebox{-1}[1]{
    \includegraphics[width=\linewidth]{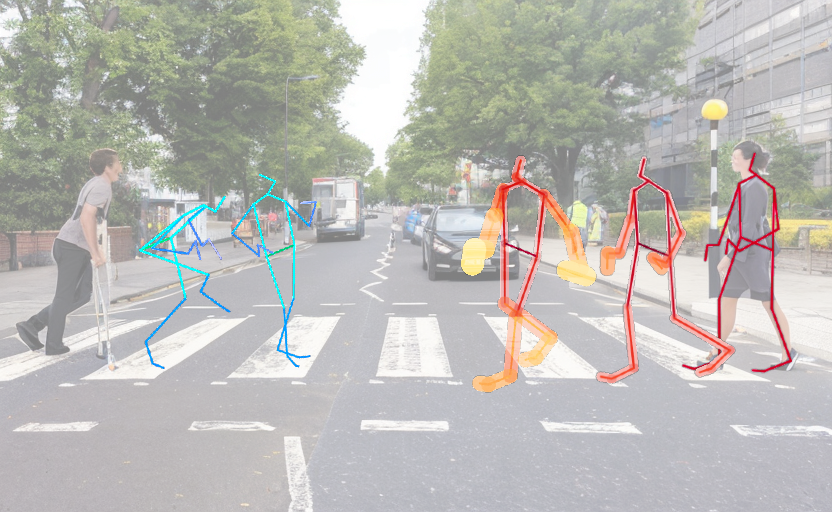}}
    \caption{We propose to model two kinds of uncertainty: 1) Aleatoric uncertainty, highlighting the inherent temporal evolution with lighter colors and thicker bones over time, illustrated by the left person; 2) Epistemic uncertainty, to detect non valid, out-of-distribution forecast poses due to unseen scenarios in training, exemplified by the right person.
  }
  \vspace{-5pt}
  \label{fig:pull_figure}
\end{figure}

Forecasting human poses is a difficult task with multiple challenges to solve: it mixes both spatial and temporal reasoning, with a huge variability in scenarios; and human behavior is difficult to predict, as it changes in dynamic and multi-modal ways to react to its environment.
To guarantee safe interactions with humans, robots should not only predict human motions, but also identify scenarios in which they are uncertain~\cite{kendall2017uncertainties, bertoni2019monoloco, liu2020simple_g, van2020duq}, and act accordingly.
As an example, \cref{fig:pull_figure} illustrates a pose forecasting scenario in autonomous driving context. Without an uncertainty measure, all the forecast poses are considered valid. However, uncertainty measures can detect unconfident outputs, so be treated with more caution.
Researchers have shown the benefits of estimating uncertainty for classification~\cite{liu2020simple_g, van2020duq} and regression tasks~\cite{kendall2017uncertainties, bertoni2019monoloco}, but how to apply it to pose forecasting is not yet studied.

In this paper, we present two solutions to capture the uncertainty of pose forecasting models from two important perspectives.
The first one deals with the aleatoric uncertainty, \ie, the irreducible intrinsic uncertainty in the data.
We reformulate the pose forecasting objective function to capture the aleatoric uncertainty.
To reduce the number of learned parameters and improve stability, we introduce uncertainty priors based on our knowledge about the uncertainty, \eg, that the uncertainty increases with time.
We then train the forecasting model with the new objective function.
This allows the model to focus its capacity to learn forecasting at shorter time horizons, where uncertainty is lower and learning is more meaningful, compared to longer ones that are intrinsically harder and uncertain to forecast. 
We apply our proposed uncertainty method to several models from the literature and evaluate on three well-known datasets (Human3.6M~\cite{h36m}, AMASS~\cite{amass2019}, 3DPW~\cite{3dpw}) and achieve up to 25\% improvements in forecasting at short horizons, with no loss on longer horizons.

The second one is about epistemic uncertainty which shows the model's lack of knowledge. To this end, we define a model-agnostic uncertainty metric to reflect the reliability and certainness of pose forecasting models in real-world scenarios, where the ground truth is absent for accuracy calculation.
Unlike previous methods which require accessing model~\cite{gal2016dropout} (\ie, white-box methods) or are specific to certain models~\cite{van2020duq}, our approach does not require access to the model (\ie, black-box approach) and is model-agnostic.
Since there is no label for motions, we train a deep clustering network to learn the distribution of common poses and measure the dissimilarity between the predictions' embeddings and cluster centers.
We achieve better performance in detecting out-of-distribution forecast poses using our epistemic uncertainty metric than other approaches from the literature.

Lastly, it is important to acknowledge that the field of pose forecasting is rapidly advancing, thanks to the significant interest from researchers and practitioners.
However, this happens at the cost of unfair and non-unified evaluations.
All current works use disparate metrics and dataset setups to report their results, leading to ambiguities and errors in interpretation.
In an effort to mitigate these discrepancies, we release an open-source library for human pose forecasting named \textit{UnPOSed}\footnote{\href{https://github.com/vita-epfl/UnPOSed}{https://github.com/vita-epfl/UnPOSed}}. This includes our re-implementations of over 10 models, processing codes for 3 widely-used datasets and 6 metrics, all implemented and tested in a standardized way, in order to ease the implementation of new ideas and promote research in this field.
To summarize our contributions:
\begin{itemize}
    \item We propose a method for incorporating priors to estimate the aleatoric uncertainty in human pose forecasting and demonstrate its efficacy in improving several state-of-the-art models on multiple datasets;
    \item We propose a model-agnostic metric of quantifying epistemic uncertainty to evaluate models in unseen situations, outperforming previous methods;
    \item We develop and publicly release an open-source library for human pose forecasting.
\end{itemize}

\section{Related works}

\textbf{Human pose forecasting:} 
While the literature has extensively examined the forecasting of a sequence of future center positions at a coarse-grained level~\cite{liu2023sim, saadatnejad2022sattack, bahari2022sattack} or a sequence of bounding boxes~\cite{bouhsain2020pedestrian, saadatnejad2022pedestrian}, 
our focus in this work is on a more fine-grained forecasting \ie, pose. 
Additionally, we limit our focus to the observation sequence alone, rather than incorporating context information~\cite{hassan2021stochastic}, social interactions~\cite{adeli2020socially}, action class~\cite{cai2021unified} or global movements~\cite{parsaeifard2021decoupled}.
Many approaches have been proposed for human pose forecasting, with some using feed-forward networks~\cite{li2018convolutional} and many others using Recurrent Neural Networks (RNNs) to capture temporal dependencies~\cite{jain2016structural, martinez2017human}. To better capture spatial dependencies of body poses, Graph Convolutional Networks (GCNs) have been utilized~\cite{mao2019ltd, mao2020history}, along with separating temporal and spatial convolution blocks and using trainable adjacency matrices~\cite{sofianos2021stsgcn}. Attention-based approaches have also gained interest for modeling human motion, showing improvement with a spatio-temporal self-attention module~\cite{mao2020history}. More recently, forecasting in multiple stages~\cite{ma2022progressively} and a diffusion model with a transformer-based architecture~\cite{saadatnejad2023diffusion} have been proposed.

We can categorize all previous works into stochastic and deterministic models. Stochastic models~\cite{yuan2020dlow, aliakbarian2020stochastic, salzmann2022motron, ma2022multiobjective, mao2021generating, saadatnejad2023diffusion, nikdel2023dmmgan} can give diverse predictions but we mainly focus on deterministic models~\cite{mao2020history,ma2022progressively,sofianos2021stsgcn,dang2021msr} as they provide more accurate predictions which is crucial for robotics applications. Given the growing interest in this field, we believe that greater attention should be paid to uncertainty estimation in this task.

\textbf{Uncertainty in pose forecasting:}
Knowing when a model does not know, \ie, uncertain, is important to improve trustworthiness and safety~\cite{mcallister2017concrete}.
Traditionally, uncertainty in deep learning is divided into data (aleatoric) and model (epistemic) uncertainty~\cite{kendall2017uncertainties}.
The aleatoric originates from the intrinsic noise and inherent uncertainty of data and cannot be reduced by improving the model, while the epistemic uncertainty shows the model’s weakness in recognizing the underlying structure of the data and can be reduced by enhancing the network architecture or increasing data. 
Many methods have been proposed to estimate and utilize these types of uncertainty in various tasks, including image classification~\cite{gal2016dropout}, semantic segmentation~\cite{kendall2015bayesian}, and natural language processing~\cite{xiao2019quantifying}.
It has also been explored in pose estimation from images and videos~\cite{9878469, 8206335}, visual navigation and trajectory forecasting tasks~\cite{9811776, 8794282} but not yet studied in human pose forecasting which includes spatio-temporal relationships modeling. We will show how modeling the uncertainty can improve accuracy.

Moreover, it is important to measure the epistemic uncertainty of models intended for real-world applications.
Bayesian Neural Networks (BNNs) have conventionally been used to formulate uncertainty by defining probability distributions over the model parameters~\cite{neal2012bayesian}. However, the intractability of these distributions has led to the development of alternative approaches to approximate Bayesian inference for uncertainty estimation.
One widely used method is Variational Inference \cite{graves2011practical, blundell2015weight}, which is valued for its scalability. A notable example is Monte Carlo (MC) dropout~\cite{gal2016dropout}, which involves applying dropout~\cite{srivastava2014dropout} at inference time to model the parameters of the network as a mixture of multivariate Gaussian distributions with small variances. However, those methods are not model-agnostic. Another approach, known as calibration~\cite{guo2017calibration}, requires the model to provide probabilities, but deep neural networks have been shown to be poorly calibrated.
One way to evaluate model reliability is by measuring the distance between a new sample and the training samples using a deep deterministic network, a technique proven effective in image classification~\cite{liu2020simple_g,van2020duq}. 
However, this approach measures the uncertainty for their own model and is not applicable to measuring the uncertainty of different models.
In contrast, Deep Ensembles \cite{lakshminarayanan2017simple} can measure the uncertainty of different models by training multiple neural networks independently and averaging their outputs at inference time. Nevertheless, this method can be computationally expensive and slow.
In this study, we concentrate on the model's output and define epistemic uncertainty as the extent to which the model's forecasts align with the training distribution, providing a black-box uncertainty measurement of pose forecasting models.

\section{Aleatoric uncertainty in pose forecasting}

Pose forecasting models usually take as input a sequence $x$ of 3D human poses with $J$ joints in $O$ observation time frames, and predict another sequence $\hat{y}$ of 3D poses to forecast its future $y$ in the next $T$ time frames.
In addition to this, we want a model to estimate its aleatoric uncertainty $u$ along with the predicted poses $\hat{y}$, to indicate how reliable these can be.

For this, we model the probability distribution of the error, \ie, the euclidean distance between ground truths $y$ and forecasts $\hat{y}$, with an exponential distribution following~\cite{bertoni2019monoloco}:
\begin{equation}
    \norm{y-\hat{y}}_2 
    \sim 
    \mathrm{Exp}(\alpha),
\end{equation}
where $\alpha$ is the distribution parameter to be selected.
Its log-likelihood therefore writes
\begin{equation}
    \ln{p(\norm{y-\hat{y}}_2)} = \ln{\alpha} - \alpha \norm{y-\hat{y}}_2.
\end{equation}
We then define the aleatoric uncertainty as $u := - \ln{\alpha}$, and set it as a learnable parameter for the model.
When training the model with maximum likelihood estimation, the loss function $\mathcal{L}$ to minimize is then given by
\begin{align}
    \mathcal{L}(y, \hat{y}, u) = -\ln{p(\norm{y-\hat{y}}_2)} = e^{-u}\norm{y-\hat{y}}_2 + u.
\end{align}

We consider pose forecasting as a multi-task learning problem with task-dependant uncertainty, \ie, independent of the input sequences $x$.
There are several ways to define tasks in this manner, \eg, by separating them based on time frames, joints, actions (if the datasets provide them), or any other combination of them.
In the following, we consider dividing tasks based on time and joints\footnote{Extending the formulation to other task definitions is straightforward.}.
In this case, for each future time frame $t$ and joint $j$, the model predicts an uncertainty estimate $u^{j}_{t}$ associated with its 3D joint forecasts $\hat{y}^{j}_{t}$.
This formulation yields the corresponding loss function:
\begin{equation}
\label{eq:final_loss}
    \mathcal{L}_{total}(y, \hat{y}, u) = \sum_{\substack{t=1 \dots T\\j=1 \dots J}} e^{-u^{j}_{t}} \norm{y^{j}_{t}-\hat{y}^{j}_{t}}_2 + u^{j}_{t},
\end{equation}
where $T$ refers to the number of prediction frames and $J$ is the number of joints.

Since the loss function (\cref{eq:final_loss}) weighs the error $\norm{y^{j}_{t}-\hat{y}^{j}_{t}}_2$ based on the aleatoric uncertainty $e^{-u^{j}_{t}}$, it forces the model to focus its capacity to points with lower aleatoric uncertainty.
In particular, we expect short time horizons to have lower uncertainty, and therefore to present better improvements than longer ones.

Unfortunately, learning all aleatoric uncertainty values $u^{j}_{t}$ independently leads to an unstable training.
To address this issue, we introduce uncertainty priors $F$, in order to inject knowledge about the aleatoric uncertainty pattern and stabilize the training.
For this, we choose a family $F$ of functions parameterized by a given number of parameters $\theta$.
Instead of learning all uncertainty values $u^{j}_{t}$ independently, the model now only learns $\theta$, which can be chosen to be of a smaller size so as to ease the training.
With a learned $\theta^*$, the uncertainty values $u^{j}_{t}$ are obtained with the function $F(\theta^*)$:
\begin{equation}
    u^{j}_{t} = F(\theta^*)(j, t).
\end{equation}
It is noticeable that this framework generalizes the previous case (without prior) by setting $F$ to yield a separate parameter for each uncertainty value:
\begin{equation}
    u^{j}_{t} = \mathrm{Id}(\theta^*)(j, t) = \theta^{j}_t.
\end{equation}

Intuitively, the more parameters $F$ has, the more scenarios it can represent, but at the cost of stability.
We, therefore, compare several choices for $F$, with variable numbers of learnable parameters as different trade-offs between ease of learning and representation power.
We select three functions that constrain the temporal evolution of aleatoric uncertainty, independently for each joint.
We select functions with a logarithmic shape due to the observed exponential pattern in error evolution over time.
The first one, $\mathrm{Sig}_3$, is a sigmoid function used to ensure that uncertainty only increases with time, and has three parameters per joint to control this pattern:
\begin{equation}
    u^{j}_{t} = \mathrm{Sig}_3(\theta)(j,t) = \frac{\theta^{j}_2}{1+e^{-\theta^{j}_0(t-\theta^{j}_1)}}.
    \label{eq:sig3}
\end{equation}
Then we leverage $\mathrm{Sig}_5$, which is a generalized version of the sigmoid function~\cite{ricketts1999five} with 5 parameters per joint:
\begin{equation}
    u^{j}_{t} = \mathrm{Sig}_5(\theta)(j,t) = \theta^{j}_0 + \frac{\theta^{j}_1}{1 + a b + (1 - a) c},
    \label{eq:sig5}
\end{equation}
where the terms $a$, $b$ and $c$ are defined by
\begin{align}
    a = \frac{1}{1 + e^{-\frac{2 \ \theta^{j}_2 \ \theta^{j}_4}{|\theta^{j}_2 + \theta^{j}_4|}  (\theta^{j}_3 - t)}}, 
    b = e^{\theta^{j}_2  (\theta^{j}_3 - t)}, 
    c = e^{\theta^{j}_4  (\theta^{j}_3 - t)}.
\end{align}
Note that optimization with $\mathrm{Sig}_5$ can converge to all uncertainty coefficients learnable with $\mathrm{Sig}_3$, but also additional values, benefiting from its strictly larger output space.

We also compare with a more generic polynomial function $\mathrm{Poly}_d$ of degree $d$, which has $d+1$ learnable parameters per joint and constrain the uncertainty less:
\begin{equation}
    u^{j}_{t} = \mathrm{Poly}_d(\theta)(j,t) = \theta^{j}_0 + \theta^{j}_1 t + \theta^{j}_2 t^2 + ... + \theta^{j}_d t^d.
\end{equation}

\section{Epistemic uncertainty in pose forecasting}
Now, we address the epistemic uncertainty to capture the model's uncertainty due to the lack of knowledge.
We want to quantify the intuition that the models with predicted motions dissimilar to the training distribution in the latent representation are less reliable and, therefore, should be treated with caution.
Notably, our aim in this section is not to improve accuracy but rather to measure uncertainties associated with pose forecasting models.

\begin{figure}[!t]
    \centering
    \includegraphics[width=0.9\linewidth]{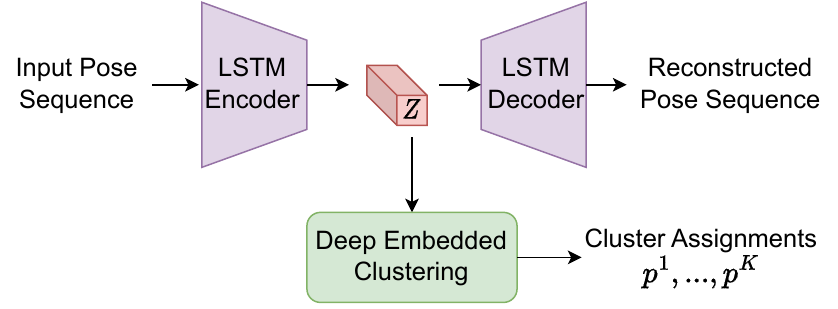}
    \caption{The motion is encoded into a well-clustered representation space $Z$ by our LSTM encoder-decoder. The probabilities of the cluster assignments are provided by our deep embedded clustering on that space to estimate the epistemic uncertainty.
    }
    \label{fig:network}
\end{figure}

We improve upon existing literature of uncertainty quantification by introducing temporal modeling and clustering in epistemic uncertainty. Specifically, we employ an LSTM-based autoencoder (Fig. \ref{fig:network}) due to its proficient capability to encode spatio-temporal dependencies and learn potent latent representations.
We then rely on clustering on that space as there are no predefined motion classes.

In the next parts, we first explain how to estimate the number of motion clusters $K$ and train the deep clustering. We then illustrate how to measure the epistemic uncertainty.

\subsection{Determining the number of motion clusters}

Determining \textit{K}, the number of clusters, is essential since it corresponds to the diversity of motions in the training dataset. An optimal \textit{K}, therefore, captures the diversity in the training dataset while also reducing the time complexity of our subsequent algorithms.

We first train an LSTM auto-encoder (\cref{fig:network}) to learn low dimensional embeddings $Z$ by minimizing the reconstruction loss $\mathcal{L}_{recons}$ over the training dataset. 
We then follow DED~\cite{wang2018ded} which uses t-SNE~\cite{JMLR:v9:vandermaaten08a} to reduce $Z$ to a 2-dimensional feature vector $z'$. 
Subsequently, local density $\rho_i$ and delta $\delta_i$ for each data point are calculated:
\begin{equation}
    \rho_i = \sum_{j} \chi(d_{ij} - d_c), \;\;\;\;
    \delta_i = \min_{j:\rho_j>\rho_i} d_{ij},
\end{equation}
where $\chi(.) = 1$ if $. < 0$ else $\chi(.) = 0$, $d_{ij}$ is the distance between $z'_i$ and $z'_j$, and $d_c$ is the cut-off distance. 
We then define $\gamma_i = \rho_i . \delta_i$ similar to ~\cite{wang2016automaticcluster}.
A larger $\gamma_i$ corresponds to a greater likelihood of being chosen as a cluster center; however, the number of clusters still remains a hyperparameter.
We fully automate it by defining $r_i$ as the gap between two $\gamma_i$ and $\gamma_{i+1}$ values (where $\gamma_{i+1} < \gamma_i $):
\begin{equation}
    r_i = \cfrac{\gamma_{i}}{\gamma_{i+1}}, i \in[1, N - 1].
\end{equation}
We set $K = argmax(r_i)$ since $\gamma_K$ represents the largest shift in likelihood of a sample being a cluster itself.

\subsection{Deep embedded clustering}
\label{sec:deepembeddedclustering}
Having identified the number of clusters, we now learn the optimal deep clustering of our embedding. 
We initialize the cluster centers $\{\mu^k\}_{k=1}^K$ using the K-means algorithm on the feature space. We then minimize the clustering loss $\mathcal{L}_{cluster}$ as defined in DEC~\cite{xie2016dec} jointly with the reconstruction loss in order to learn the latent representation as well as clustering. We incorporated the reconstruction loss into the loss function to act as a regularizer and prevent the collapse of the network parameters. The loss function is defined as:
\begin{equation}
\mathcal{L} = \mathcal{L}_{cluster} + \lambda \mathcal{L}_{recons},
\label{eq:clustering_loss}
\end{equation}
where $\lambda$ is the regularization coefficient. Finally, when the loss is converged, we fine-tune the trained network using the cross-entropy loss on the derived class labels in order to make clusters more compact.

\subsection{Estimating epistemic uncertainty}
Now, we estimate the epistemic uncertainty of a given forecasting model. Specifically, for each example, denote the probability of assignment to the $k$th cluster by $p^{k}$. The epistemic uncertainty is then calculated as follows:
\begin{equation}
    EpU = \frac{1}{N}\sum_{i=1}^{N}\textrm{entropy}(p^{1}_i, \dots, p^{K}_i),
\label{eq:oodu}
\end{equation}
where $N$ is the size of the dataset. In other words, a model that does not generate outputs close to the motion clusters is considered uncertain.

\section{Experiments}

\subsection{Datasets and Metrics}

\textbf{Human3.6M}~\cite{h36m} contains 3.6 million body poses. It comprises 15 complex action categories, each one performed by seven actors individually.
The validation set is subject-11, the test set is subject-5, and all the remaining five subjects are training samples.
The original 3D pose skeletons in the dataset consist of 32 joints.
Similar to previous works, we have  10/50 observation frames, 25 forecast frames down-sampled to 25 fps, with the subset of 22 joints to represent the human pose.

\textbf{AMASS} (The Archive of Motion Capture as Surface Shapes) ~\cite{amass2019} unifies 18 motion capture datasets totaling 13,944 motion sequences from 460 subjects performing a variety of actions.
We use 50 observation frames down-sampled to 25 fps with 18 joints, similar to previous works.

\textbf{3DPW} (3D Poses in the Wild)~\cite{3dpw} is the first dataset with accurate 3D poses in the wild. It contains 60 video sequences taken from a moving phone camera. Each pose is described as an 18-joint skeleton with 3D coordinates similar to AMASS dataset.  
We use the official instructions to obtain training, validation and test sets.

We measure the accuracy in terms of MPJPE (Mean Per Joint Position Error) in millimeters (mm) per frame
and in terms of A-MPJPE as the average for all frames when needed.
We also report EpU as defined in \cref{eq:oodu}.

\subsection{Baselines}
We apply our approach to several recent methods that are open-source~\cite{mao2020history, ma2022progressively, sofianos2021stsgcn} and compare the performances of with and without the incorporation of our approach.
Note that we follow their own training setup in which some use 10 frames of observation~\cite{sofianos2021stsgcn,ma2022progressively,dang2021msr} and the rest 50 frames of observation~\cite{mao2020history, martinez2017human, li2018convolutional,mao2019ltd}.
We report the results obtained from the pretrained model of deterministic STARS*~\cite{xu22stars} as documented on their GitHub page.
We also consider \textit{Zero-Vel}, a simple and competitive baseline~\cite{martinez2017human}, that forecasts all future poses by outputting the last observed pose.

\begin{figure}[!t]
    \centering
    \includegraphics[width=0.75\linewidth]{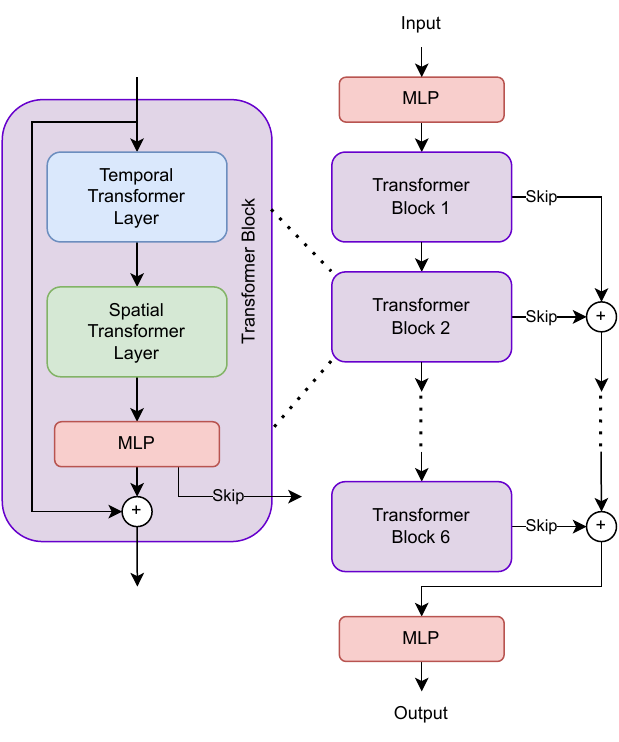}
    \caption{ST-Trans consists of 2 MLP layers and 6 Transformer Blocks with skip connections. Each Transformer Block contains two cascaded temporal and spatial transformers to capture the spatio-temporal features of data.}
    \label{fig:st_trans}
\end{figure}

Inspired by the common trend to treat sequences with Transformers, we have designed our own simple transformer-based architecture referred to as \textit{ST-Trans}. We followed the best practices proposed in~\cite{tashiro2021csdi} and adapted their design elements to the task of pose forecasting.
As depicted in \cref{fig:st_trans}, it is composed of several identical residual layers, each layer consists of a spatial and a temporal transformer encoder to learn the spatio-temporal dynamics of data utilizing the attention mechanism.

\subsection{Aleatoric uncertainty}

\begin{table*}[!t]
    \centering
    \begin{tabular}{lcccccccc}
        \toprule
        Model & $80\,\mathrm{ms}$ & $160\,\mathrm{ms}$ & $320\,\mathrm{ms}$ & $400\,\mathrm{ms}$ & $560\,\mathrm{ms}$ & $720\,\mathrm{ms}$ & $880\,\mathrm{ms}$ & $1000\,\mathrm{ms}$  \\
        \midrule
        Zero-Vel~\cite{martinez2017human} & 23.8 & 44.4 & 76.1 & 88.2 & 107.4 & 121.6 & 131.6 & 136.6 \\
        Res. Sup.~\cite{martinez2017human} & 25.0 & 46.2 & 77.0 & 88.3 & 106.3 & 119.4 & 130.0 & 136.6 \\         ConvSeq2Seq~\cite{li2018convolutional} & 16.6 & 33.3 & 61.4 & 72.7 & 90.7 & 104.7 & 116.7 & 124.2 \\ 
        LTD-50-25~\cite{mao2019ltd} & 12.2 & 25.4 & 50.7 & 61.5 & 79.6 & 93.6 & 105.2 & 112.4 \\
        MSR-GCN~\cite{dang2021msr} & 12.0 & 25.2 & 50.4 & 61.4 & 80.0 & 93.9 & 105.5 & 112.9 \\
        STARS*~\cite{xu22stars} & 12.0 & 24.6 & 49.5 & 60.5 & 78.6 & 92.6 & 104.3 & 111.9 \\
        \midrule
        STS-GCN~\cite{sofianos2021stsgcn} & 17.7 & 33.9 & 56.3 & 67.5 & 85.1 & 99.4 & 109.9 & 117.0 \\
        STS-GCN + pUAL (ours) & 13.2 & 27.1 & 54.7 & 66.2 & 84.5 & 97.9 & 109.3  & 115.7  \\
        gain & \textbf{25.4\,\%} & 20.1\,\% & 2.8\,\% & 1.9\,\% & 0.7\,\% & 1.5\,\% & 0.5\,\% & 1.1\,\% \\
        \midrule
        HRI*~\cite{mao2020history} & 12.7 & 26.1 & 51.5 & 62.6 & 80.8 & 95.1 & 106.8 & 113.8 \\
        HRI* + pUAL (ours) & 11.6 & 25.3 & 51.2 & 62.2 & 80.1 & 93.7 & 105.0 & 112.1 \\
        gain & \textbf{8.7\,\%} & 3.1\,\% & 0.6\,\% & 0.6\,\% & 0.9\,\% & 1.5\,\% & 1.7\,\% & 1.5\,\% \\
        \midrule
        PGBIG~\cite{ma2022progressively} & 10.3 & 22.6 & 46.6 & 57.5 & 76.3 & 90.9 & 102.7 & 110.0 \\
        PGBIG + pUAL (ours) & 9.6 & 21.7 & 46.0 & 57.1 & 75.9 & 90.3 & 102.1 & 109.5\\
        gain & \textbf{6.8\,\%} & 4.0\,\% & 1.3\,\% & 0.7\,\% & 0.5\,\% & 0.7\,\% & 0.6\,\% & 0.5\,\% \\
        \midrule
        ST-Trans & 13.0 & 27.0 & 52.6 & 63.2 & 80.3 & 93.6 & 104.7 & 111.6 \\
        ST-Trans + pUAL (ours) & 10.4 & 23.4 & 48.4 & 59.2 & 77.0 & 90.7 & 101.9 & 109.3 \\
        gain & \textbf{20.0\,\%} & 13.3\,\% & 8.0\,\% & 6.3\,\% & 4.1\,\% & 3.1\,\% & 2.7\,\% & 2.1\,\% \\
        \bottomrule
    \end{tabular}
    \caption{Comparison of our method on Human3.6M~\cite{h36m} in MPJPE ($\mathrm{mm}$) at different prediction horizons. +pUAL refers to models where aleatoric uncertainty is modeled.}
    \label{tab:h36_alu}
\end{table*}

\begin{figure*}[!t]
  \begin{subfigure}[b]{0.33\linewidth}
    \centering
    \includegraphics[width=0.82\linewidth]{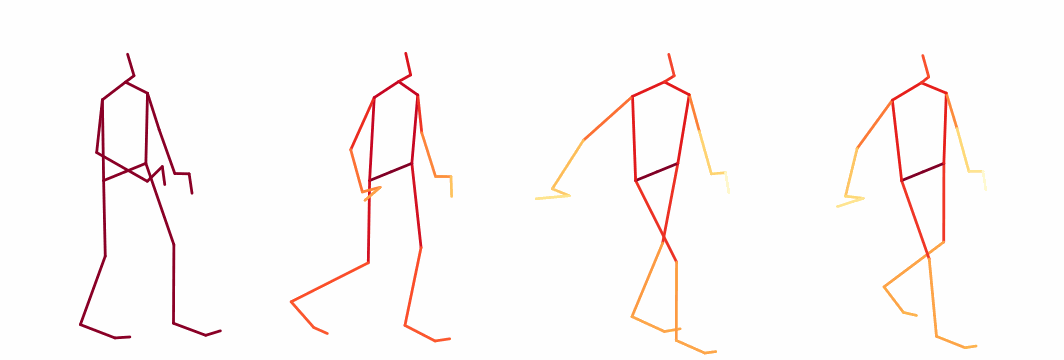} 
    \caption{Walking} 
    \label{fig7:a} 
  \end{subfigure}
  \begin{subfigure}[b]{0.33\linewidth}
    \centering
    \includegraphics[width=0.82\linewidth]{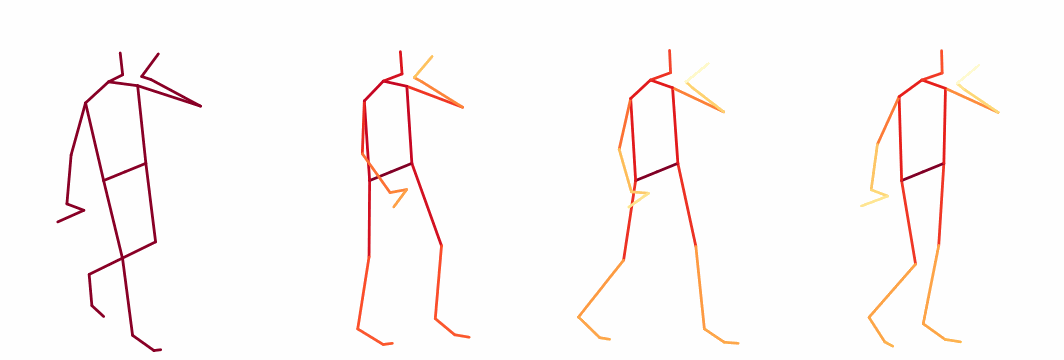} 
    \caption{Phoning} 
    \label{fig7:c} 
  \end{subfigure}
  \begin{subfigure}[b]{0.33\linewidth}
    \centering
    \includegraphics[width=0.82\linewidth]{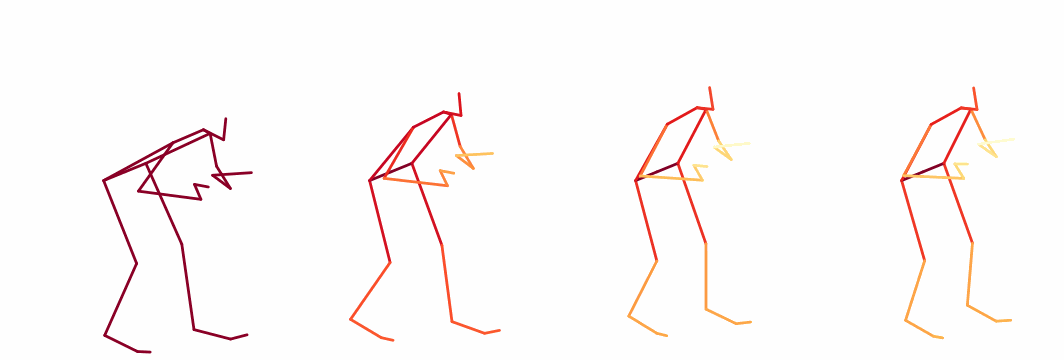}
    \caption{Taking Photo} 
    \label{fig7:d} 
  \end{subfigure} 
  \caption{Qualitative forecast poses on Human3.6M~\cite{h36m} depicting different actions over time. For each action, time progresses from left to right. Higher aleatoric uncertainty is shown with a lighter color. Uncertainty of any bone is considered as its outer joint's uncertainty assuming the hip is the body center. We observe that the estimated uncertainty increases over time, with joints farther away from the body center associated with higher uncertainties.}
  \label{fig:qual}
\end{figure*}

\begin{table}[!t]
    \centering
    \resizebox{\columnwidth}{!}{
    \begin{tabular}{lccccccccc}
        \toprule
        \multirow{2}[3]{*}{Model} & \multicolumn{4}{c}{AMASS} & \multicolumn{4}{c}{3DPW} \\
    	\cmidrule(lr){2-5} \cmidrule(lr){6-9}
        & $160\,\mathrm{ms}$ & $400\,\mathrm{ms}$ & $720\,\mathrm{ms}$ & $1000\,\mathrm{ms}$ & $160\,\mathrm{ms}$ & $400\,\mathrm{ms}$ & $720\,\mathrm{ms}$ & $1000\,\mathrm{ms}$ \\
        \midrule
        Zero-Vel~\cite{martinez2017human} & 56.4 & 111.7 & 135.1 & 119.4 & 41.8 & 79.9 & 100.5 & 101.3 \\
        ConvSeq2Seq~\cite{li2018convolutional} & 36.9 & 67.6 & 87.0 & 93.5 & 32.9 & 58.8 & 77.0 & 87.8 \\
        LTD-10-25~\cite{mao2019ltd} & 20.7 & 45.3 & 65.7 & 75.2 & 23.2 & 46.6 & 65.8 & 75.5 \\
        \midrule
        STS-GCN~\cite{sofianos2021stsgcn} & 20.7 & 43.1 & 59.2 & 68.7 & 20.8 & 40.3 & 55.0 & 62.4 \\
        STS-GCN + pUAL & 20.4 & 42.4 & 59.1 & 68.1 & 20.5 & 40.0 & 54.8 & 62.2 \\
        \midrule
        HRI~\cite{mao2020history} & 20.7 & 42.0 & 58.6 & 67.2 & 22.8 & 45.0 & 62.9 & 72.5 \\
        HRI + pUAL & 19.9 & 41.4 & 58.1 & 66.5 & 22.2 & 44.6 & 62.4 & 72.2 \\
        \midrule
        ST-Trans & 21.3 & 42.5 & 58.3 & 66.6 & 24.5 & 47.4 & 64.6 & 73.8 \\
        ST-Trans + pUAL & 18.3 & 39.7 & 56.5 & 66.7 & 22.3 & 45.7 & 63.6 & 73.2 \\
        \bottomrule
    \end{tabular}
    }
    \caption{Comparison of our proposed method on AMASS~\cite{amass2019} and 3DPW~\cite{3dpw} in MPJPE ($\mathrm{mm}$) at different prediction horizons. +pUAL refers to models where aleatoric uncertainty is modeled. The models were trained on AMASS.}
    \label{tab:amass_and_3dpw}
\end{table}

We first show the impact of aleatoric Uncertainty-Aware Loss (pUAL) with the prior $\mathrm{Sig}_5$ to several models from the literature and our ST-Trans.
\cref{tab:h36_alu} shows the overall results on Human3.6M~\cite{h36m}.
To have a fair evaluation between all models, we adapt HRI~\cite{mao2020history} to predict 25 frames in one step (denoted as HRI*).
We observe that all methods get better results when taking aleatoric uncertainty into account during learning, therefore confirming the need for aleatoric uncertainty estimation.
It is noticeable that pUAL gives better improvements for shorter prediction horizons, \eg, up to 25.4\,\% and 20.1\,\% for STS-GCN~\cite{sofianos2021stsgcn} at horizons of $80\,\mathrm{ms}$ and $160\,\mathrm{ms}$, which correspond to the less uncertain time frames, where pUAL focuses training more (smaller discount in the loss function, as seen in \cref{eq:final_loss}).
At the same time, adding pUAL does not degrade the performances at longer horizons.
In the context of close human-robot interactions, this improved precision can significantly enhance the overall system performance.
Examples of predicted 3D pose sequences using pUAL are depicted in \cref{fig:qual}, and show that the estimated uncertainty increases over time, with joints farther away from the body center associated with higher uncertainties.
Moreover, we report the performances of the models on AMASS and 3DPW datasets in \cref{tab:amass_and_3dpw}.
Again, we observe that modeling aleatoric uncertainty leads to more accurate predictions, especially at shorter horizons, with improvements up to 14.1\,\% on AMASS and up to 9.0\,\% on 3DPW for ST-Trans at a horizon of $160\,\mathrm{ms}$.

\begin{figure}[!t]
    \centering
    \includegraphics[width=0.85\linewidth]{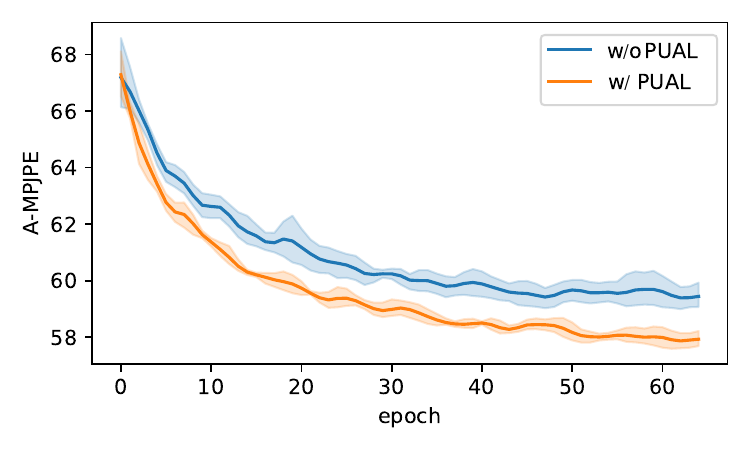}
    \caption{A-MPJPE and its standard deviation in training epochs for 5 trained models. The model with pUAL has a lower standard deviation, meaning a more stable training.}
    \label{fig:stable-training}
\end{figure}

We argue that modeling the aleatoric uncertainty leads to more stable training.
In order to demonstrate this, we conduct five separate trainings of ST-Trans and present in \cref{fig:stable-training} the average of the A-MPJPE values along with their respective standard deviations for each epoch.
The plot highlights that the model with pUAL is more stable across runs, as indicated by a lower standard deviation.
Moreover, we compute AP-MPJPE, which is the average pairwise distance of predicted motions in terms of MPJPE, and observe that it decreases from $24.2\,\mathrm{mm}$ to $20.3\,\mathrm{mm}$ when pUAL loss is added, showing again lower standard deviation in the model's output.

So far, results have been reported using the $\mathrm{Sig}_5$ uncertainty prior (\cref{eq:sig5}) to model the time and joint ($T,J$) aleatoric uncertainty.
In \cref{tab:priors}, we report the performances of other choices, and compare against using a single prior $\mathrm{Sig}_5$ for all joints (only time dependency $T$) and other priors  $\mathrm{Sig}_3$, $\mathrm{Poly}_9$.
The results show again that taking aleatoric uncertainty into account with pUAL is beneficial and that a good choice of uncertainty prior is important.
In particular, $\mathrm{Sig}_5$ performs better than using no prior for all models.
Using a prior can lead to similar aleatoric uncertainty than the unconstrained case, but with fewer learnable parameters and better stability.

\begin{table}[!t]
    \centering
    \resizebox{\columnwidth}{!}{
    \begin{tabular}{lcc|ccc}
        \toprule
        \begin{tabular}{@{}c@{}}Uncertainty\\prior (tasks)\end{tabular} & \begin{tabular}{@{}c@{}}Number of\\parameters\end{tabular} & \begin{tabular}{@{}c@{}} Standard \\ deviation \end{tabular} & ST-Trans & HRI* & STS-GCN\\
        \midrule
        None & -- & 0.643 & 111.6 & 113.8 & 117.0\\
        \midrule
        $\mathrm{Id}$ ($T,J$) & $25\cdot22$ &0.557& \textbf{109.3} & 114.6 & 115.8\\
        $\mathrm{Poly}_9$ ($T,J$) & $10\cdot22$ &0.505& 110.3 & 114.7 & 118.1\\
        $\mathrm{Sig}_5$ ($T,J$) & $5\cdot22$ &\textbf{0.496}& \textbf{109.3} & \textbf{112.1} & \textbf{115.7}\\
        $\mathrm{Sig}_3$ ($T,J$) & $3\cdot22$ &0.537& 110.3 & 113.1 & 115.9 \\
        $\mathrm{Sig}_5$ ($T$) & $5$ & 0.505 & 109.7 & 112.4 & 115.9 \\
        \bottomrule
    \end{tabular}
    }
    \caption{Comparison of different priors for aleatoric uncertainty in terms of MPJPE ($\mathrm{mm}$) at $1\,\mathrm{s}$ on Human3.6M. Lower standard deviation in training is better.}
    \label{tab:priors}
\end{table}

\subsection{Epistemic uncertainty}

Evaluating the quality of epistemic uncertainty is difficult due to the unavailability of ground truth annotations, yet important.
Our goal is to identify instances where pose forecasting is not reliable, essentially making this a binary classification problem.
Selective classification is a widely used methodology to evaluate uncertainty quality, where a classifier has the option to refrain from classifying data points if its confidence level drops below a certain threshold~\cite{el2010foundations}.
In other words, if a pose forecasting model is trained on action A and evaluated on actions A and B, a reliable measure of epistemic uncertainty should effectively distinguish between these two sets of forecasts.

\begin{figure}
    \centering
    \includegraphics[width=0.79\linewidth]{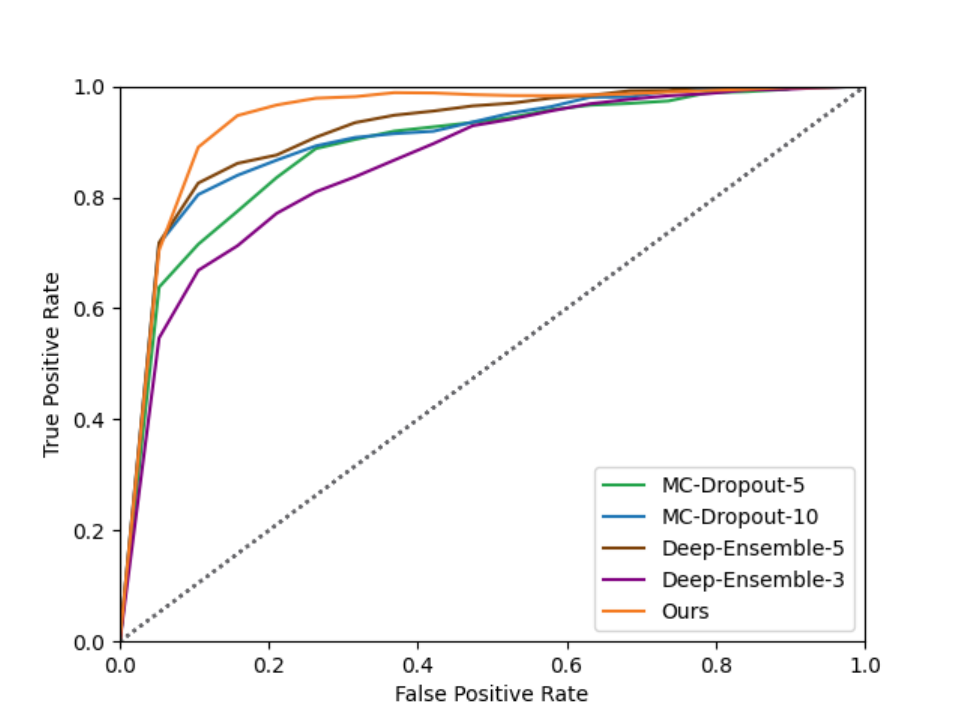}
    \caption{ROC curve for a model trained on walking-related actions and tested on both walking-related and sitting-related actions. The objective is to distinguish between these sets by utilizing uncertainty estimates.}
    \label{fig:roc}
\end{figure}

We assess the performance of our epistemic uncertainty estimation using selective classification, and measure how well actions "sitting" and "sitting down" can be separated from actions "walking" and "walking together", all from the test set of Human3.6M, based solely on the predicted uncertainty of the model.
The forecasting model and clustering are trained on Human3.6M walking-related actions, and we anticipate low uncertainty values for those actions and high uncertainty values for sitting-related actions, \ie, not encountered and significantly distinct actions. During the assessment, we compute uncertainty scores for both actions and measure the classification results for a range of thresholds. Similar to prior research~\cite{ren2019likelihood}, we utilize the AUROC metric, where a higher score is desirable and a value of $1$ indicates that all walking-related data points possess lower uncertainty than all sitting-related data points.
In \cref{tab:auroc}, we present our findings and compare them to alternative approaches, where our proposed method demonstrates higher AUROC.
The full ROC curve is in \cref{fig:roc}.
Note that our approach is model-agnostic in contrast to MC-Dropout.

Another feature of our approach is computational efficiency, which is attributed to its ability to compute in a single forward path. This is in contrast to MC-Dropout and Ensemble methods. 
We provide a comparison of the average inference latency, measured in milliseconds, between our method and other approaches in \cref{tab:auroc}. Our approach shows lower latency and only requires one training. Notably, the performance gap between our approach and other methods may increase when using more computationally expensive forecasting models.

\begin{table}[!t]
    \centering
    \begin{tabular}{lccc}
        \toprule
        Method & AUROC & Latency & Trainings \\
        \midrule
        Deep-Ensemble-3 & 0.87 & 6.28 & 3\\ 
        Deep-Ensemble-5 & 0.90 & 10.43 & 5\\ 
        MC-Dropout-5 & 0.90 & 9.57 & 1\\
        MC-Dropout-10 & 0.92 & 18.98 & 1\\
        Ours & \textbf{0.95} & \textbf{6.23} & 1\\ 
        \bottomrule
    \end{tabular}
    \caption{AUROC, inference latency ($\mathrm{ms}$) and the number of training runs for different epistemic uncertainty methods.}
    \label{tab:auroc}
\end{table}

We conducted another experiment to showcase our metric's effectiveness in out-of-distribution (OOD) motions.
By shuffling the frames' order or joints in each pose sequence of the test set, we generated OOD data. The EpU on the original test set is $0.085$, while for shuffled joints, it was observed to be $1.53$ due to the lack of correspondence with in-distribution (ID) poses. Furthermore, a high EpU value of $2.18$ was obtained for shuffled frames, highlighting the importance of frame order in generating an ID motion. 
The full table of performances in all actions can be found in the appendix.

Additionally, we report the forecasting models' performances in \Cref{tab:oodu2} in terms of A-MPJPE, along with the epistemic uncertainties EpU associated with their predictions, on both the AMASS and 3DPW datasets.
Note that the forecasting models and the clustering method were trained on the AMASS dataset.
Higher uncertainties were recorded on 3DPW as an unseen dataset while prediction errors were lower. It underscores the reliability of our uncertainty quantification approach and suggests that relying solely on a model's prediction errors may not provide a comprehensive assessment.

\begin{table}[!t]
    \centering
    \begin{tabular}{lcccccc}
        \toprule
        & \multicolumn{2}{c}{AMASS} & \multicolumn{2}{c}{3DPW} \\
        \cmidrule(lr){2-3} \cmidrule(lr){4-5}
        Model & EpU & A-MPJPE & EpU & A-MPJPE\\
        \midrule
        Zero-Vel~\cite{martinez2017human} & 0.449 & 85.72 & 0.566 & 64.44 \\
        HRI~\cite{mao2020history} & 0.351  & 43.76 & 0.463 & 43.62\\
        STS-GCN~\cite{sofianos2021stsgcn} & 0.332 & 45.49 & 0.455 & 42.60\\
        ST-Trans + pUAL &  0.336 & 35.86 & 0.439 & 40.02 \\
        \bottomrule
    \end{tabular}
    \caption{Comparison of different models in terms of A-MPJPE and EpU on AMASS and 3DPW datasets. The clustering and forecasting models were trained on AMASS.
    }
    \label{tab:oodu2}
\end{table}

\section{Conclusion}
In this paper, we focused on modeling the uncertainty of human pose forecasting.
We suggested a method for modeling aleatoric uncertainty of pose forecasting models that could make state-of-the-art models uncertainty-aware and improve their performances. We showed the effect of uncertainty priors to inject knowledge about the pattern of uncertainty.
Moreover, we measured the epistemic uncertainty of pose forecasting models by clustering poses into motion clusters, which enables us to evaluate the trustworthiness of victim models.
We made an open-source library of human pose forecasting with several models, datasets, and metrics to move toward a unified and fair evaluation.
We hope that the findings and the library will pave the way to more uncertainty-aware pose forecasting models.

\section{Acknowledgment}
The authors extend their sincere gratitude to Armin Saadat and Nima Fathi for their invaluable contributions in the project's initial phase and library development. Special thanks also go to Mohamad Asadi, Ali Rasekh, Megh Shukla, and Mohammadhossein Bahari for their helpful input.
This project has received funding from the European Union's Horizon 2020 research and innovation programme under the Marie Sklodowska-Curie grant agreement No 754354, and SNSF Sinergia Fund.


\bibliographystyle{ieee_fullname}
\bibliography{references}

\newpage

\section{Appendix}

Here, we provide supplementary materials. It comprises an experiment that explores the uncertainties of various joints and priors, and qualitative results related to aleatoric uncertainty. It also includes an evaluation of EpU across a wider range of actions, as well as a OOD motion experiment, and a discussion on motion clustering for epistemic uncertainty.

\subsection{Aleatoric uncertainty in pose forecasting}

\subsubsection{Study of different joints' aleatoric uncertainties}
\begin{figure}[!h]
\centering{
    \includegraphics[width=0.8\linewidth]{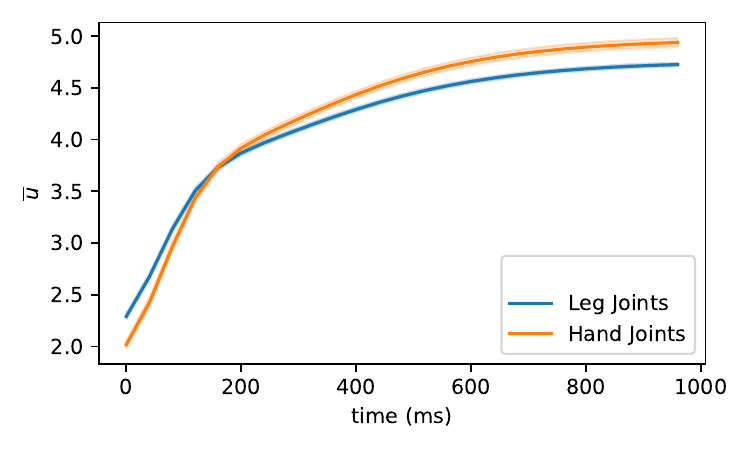}
}
\caption{Evolution of uncertainty of hands and legs over time. Hands' uncertainty is lower at short prediction horizon, but higher at longer prediction horizons.}
\label{fig:hand_leg}
\end{figure}

In the main paper, we observe that the uncertainty of joints increases over time. Another observation in our experiments is that different joints have different behaviors. For instance, \Cref{fig:hand_leg} shows that hand joints have lower uncertainties compared to leg joints in the beginning of the forecasting as hands move less and are more predictable. However, toward the end of the forecasting, hands are more unpredictable, therefore have higher uncertainties compared to legs.

\subsubsection{Study of different priors}
\begin{figure}[!h]
    \centering
    \includegraphics[width=0.8\linewidth]{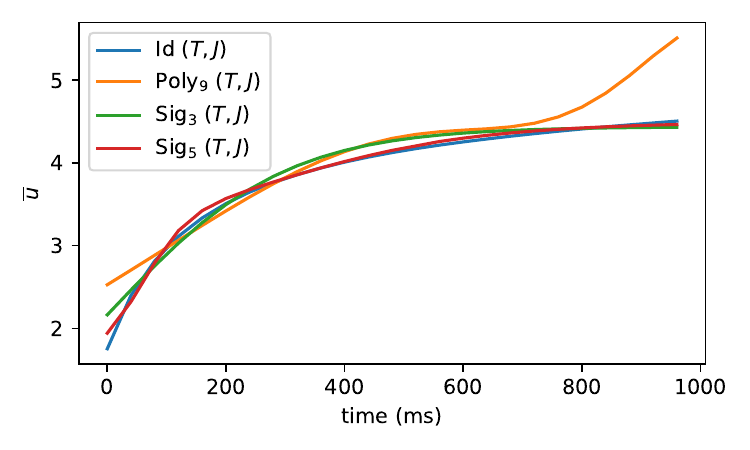}
    \caption{The values of the learned aleatoric uncertainties for different priors trained on ST-Trans.}
    \label{fig:priors}
\end{figure}

In the main paper, we observed that using a prior can lead to similar aleatoric uncertainty than the unconstrained case, but with fewer learnable parameters and better stability.
Here, we plot the  learned aleatoric uncertainty of different prior functions in \Cref{fig:priors}. We observe that all priors lead to the same general evolution over time which comes from the exponential behavior of error in time; however, ${Sig}_5$ matches the best.

\subsubsection{Qualitative results}
Examples of forecast pose sequences are depicted in \Cref{fig:q2}. We observe higher uncertainties for later time frames.

\begin{figure*}[!h]
\centering
\begin{frame}{}
  \animategraphics[loop,controls,width=0.10\linewidth]{12}{aleatoric-imgs/discussion_780/frame_}{0}{24} \qquad\qquad\qquad\qquad\qquad\qquad
  \animategraphics[loop,controls,width=0.10\linewidth]{12}{aleatoric-imgs/phoning_1650/frame_}{0}{24}\qquad\qquad\qquad\qquad\qquad\qquad
  \animategraphics[loop,controls,width=0.10\linewidth]{12}{aleatoric-imgs/purchases_2120/frame_}{0}{24} \\ \;
  \animategraphics[loop,controls,width=0.10\linewidth]{12}{aleatoric-imgs/taking_photo_3050/frame_}{0}{24} \qquad\qquad\qquad\qquad\qquad\qquad
  \animategraphics[loop,controls,width=0.10\linewidth]{12}{aleatoric-imgs/walking_0/frame_}{0}{24} \qquad\qquad\qquad\qquad\qquad\qquad
  \animategraphics[loop,controls,width=0.10\linewidth]{12}{aleatoric-imgs/walkingdog_3470/frame_}{0}{24}
\end{frame}
    \caption{Six animations showing different forecast pose sequences. Higher aleatoric uncertainty is shown with a lighter color. It is best viewed using Adobe Acrobat Reader.}
    \label{fig:q2}
\end{figure*}

\subsection{Epistemic uncertainty in pose forecasting}

\subsubsection{Evaluating EpU on OOD motions}

The full table of the OOD motions experiment of the main paper is in \Cref{tab:shuffle-supp}.

\begin{table}[!h]
\centering
\begin{tabular}{lcccccccc}
\hline
\toprule
Action        &  Normal &  Frames  & Joints \\ 
&& Shuffled & Shuffled \\ \midrule
Walking   & 0.26               & 1.35      & 2.15\\
Smoking   & 0.80               & 1.54       & 2.17\\
Posing    & 0.93               & 1.57       & 2.17\\
Directions  & 0.93               & 1.39      & 2.20\\
Greeting     & 0.81               &  1.50    & 2.17\\
Discussion     & 0.80               &  1.31  & 2.19\\
Walkingtogether     & 0.33               &  1.36     & 2.21\\
Eating     & 0.83               &  1.27  & 2.19\\
Phoning     & 0.82               &  1.56       & 2.20\\
Sitting     & 1.12               &  1.75        & 2.23\\
Waiting    & 0.82               &  1.57        & 2.15\\
Sittingdown     & 1.18               &  1.89        & 2.16\\
WalkingDog     & 0.95               &  1.53        & 2.20\\
TakingPhoto     & 1.02               &  1.47   & 2.17\\
Purchases     & 0.99               &  1.47       & 2.24\\ 
\midrule
\textbf{Average of all actions}     & \textbf{0.85}               & \textbf{1.53}    & \textbf{2.18}  \\
\bottomrule
\end{tabular}
\caption{Comparison of EpU on different categories of Human3.6M. \textit{Normal} refers to the original test set, \textit{Frames Shuffled} refers to the test set in which the frame orders in each sequence have been randomly shuffled, and \textit{Joints Shuffled} refers to randomly shuffled 3D joints in all frames.}
\label{tab:shuffle-supp}
\end{table}

\subsubsection{Evaluating EpU on other actions}
\begin{figure*}[!h]
    \centering{
    \includegraphics[width=0.32\linewidth]{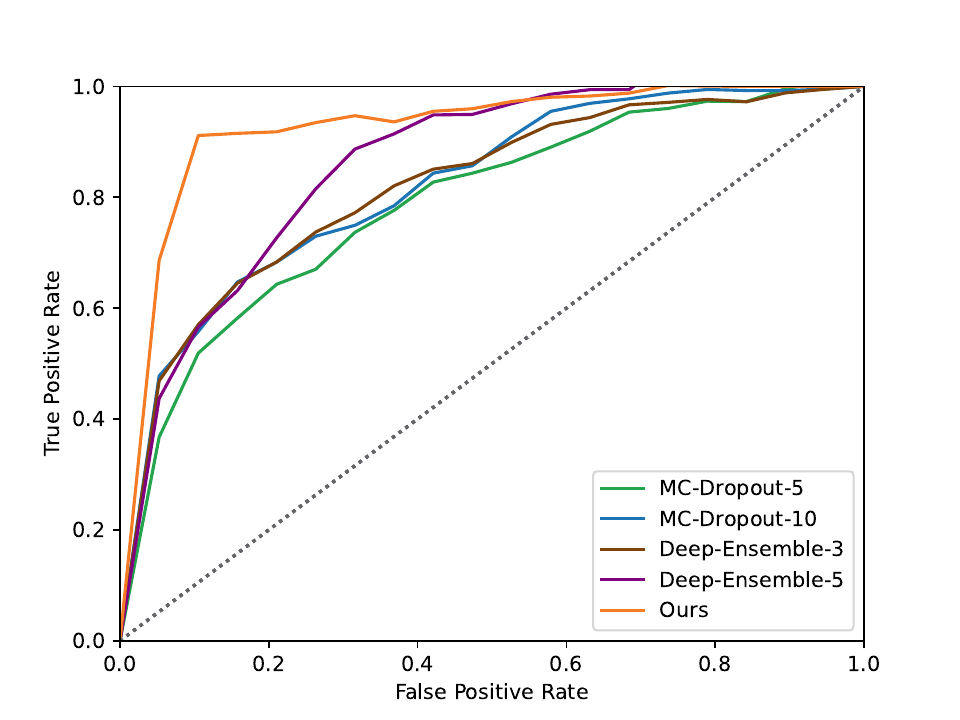}
    \includegraphics[width=0.32\linewidth]{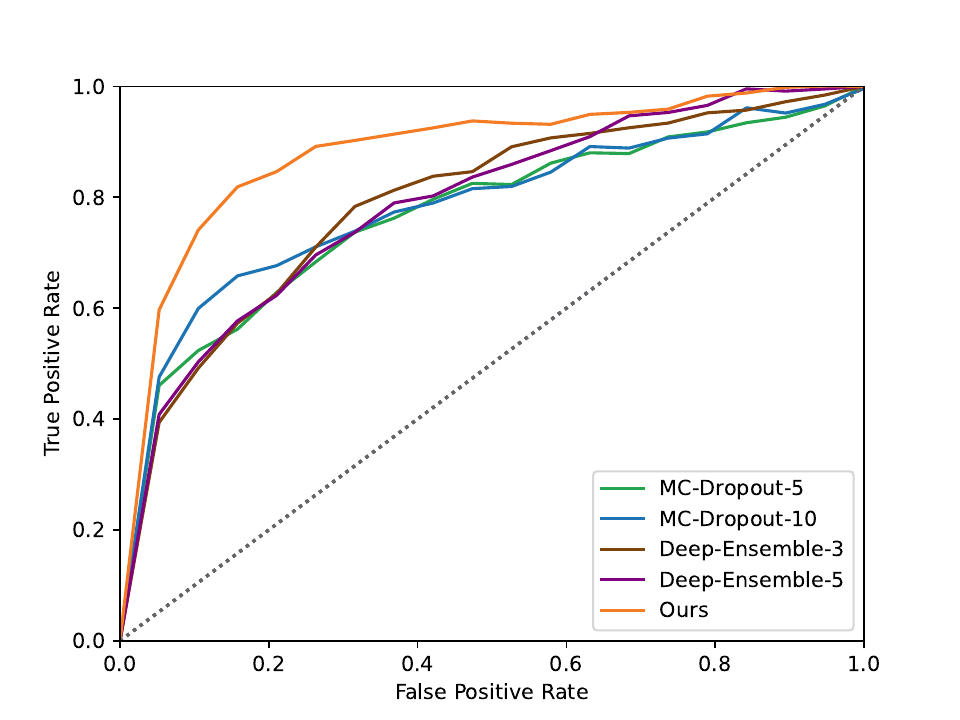}
    \includegraphics[width=0.32\linewidth]{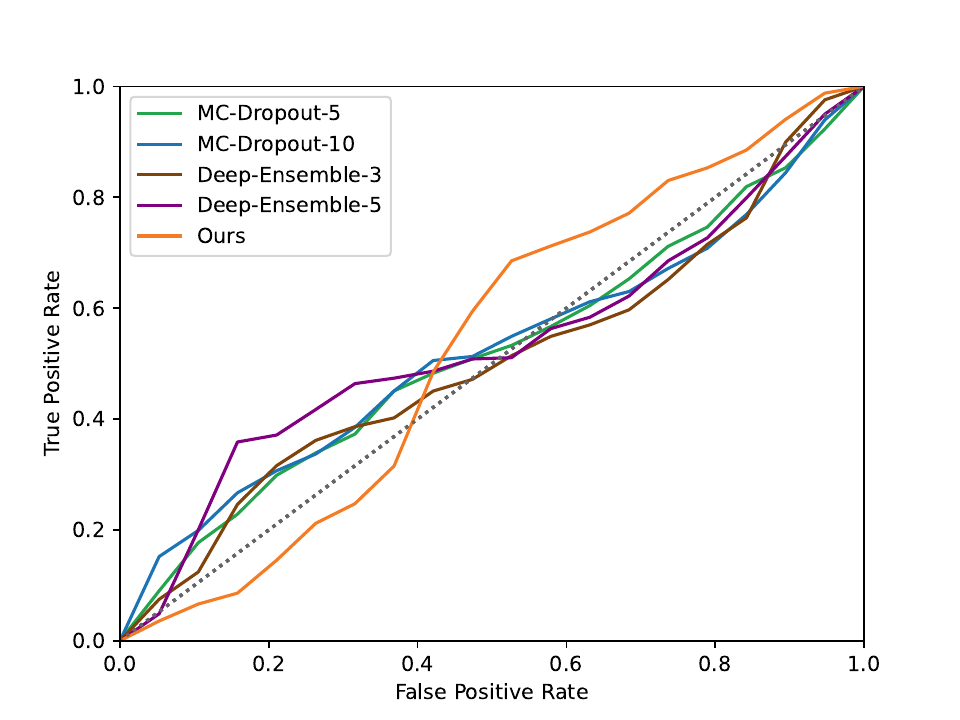} \\
    \includegraphics[width=0.32\linewidth]{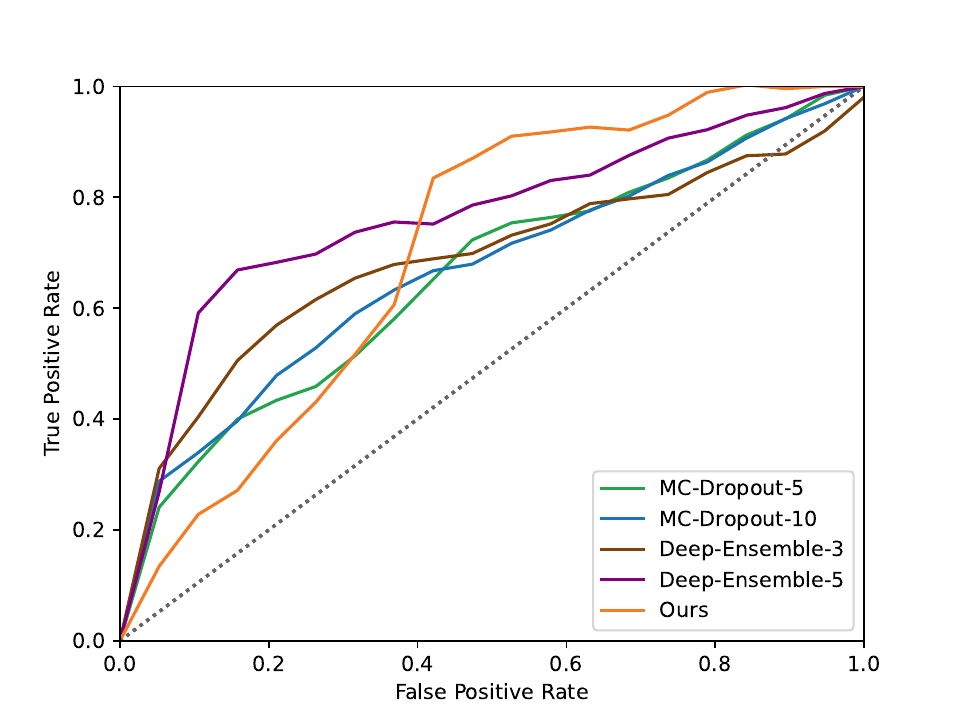}
    \includegraphics[width=0.32\linewidth]{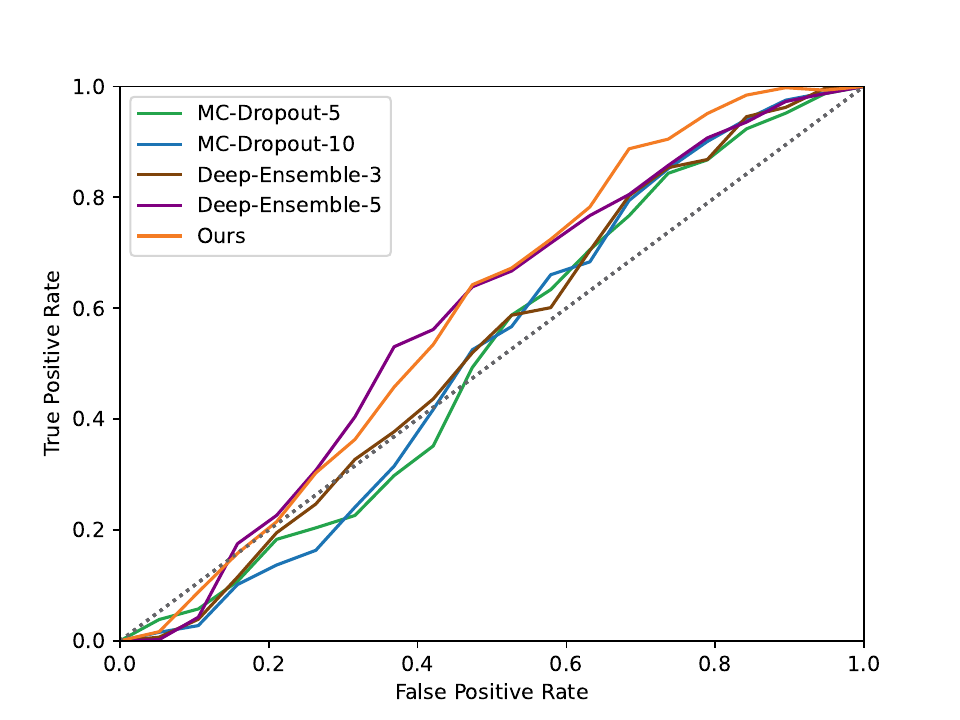}
    \includegraphics[width=0.32\linewidth]{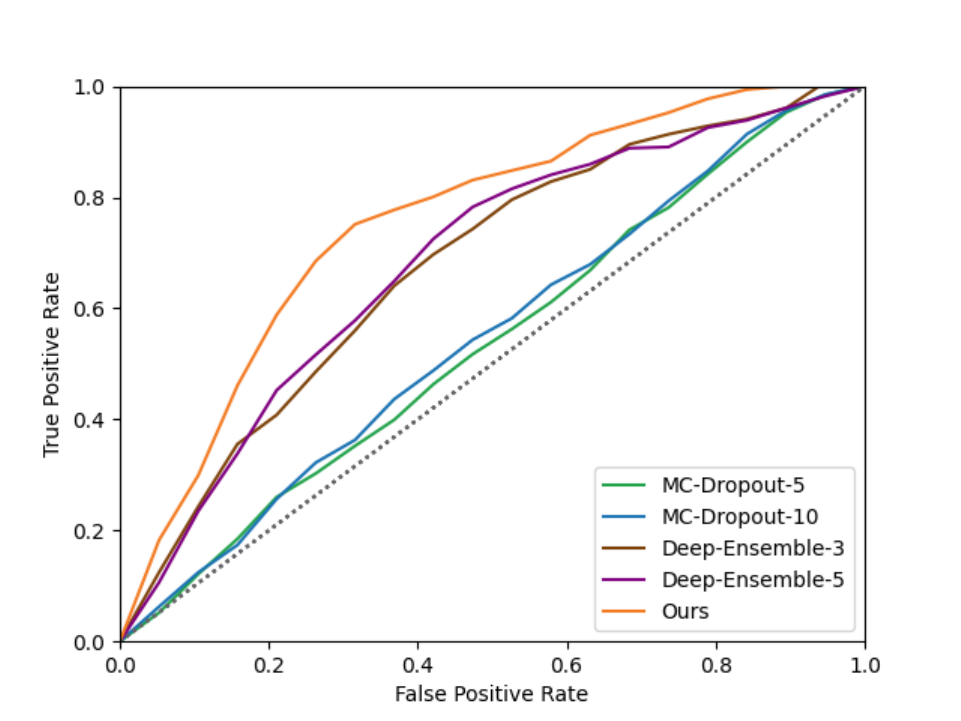}

    }
    \caption{ROC curves for a model trained on the first set of actions and tested on both first and second sets of actions. The objective is to distinguish between these sets by utilizing uncertainty estimates.  (a) Walking - Purchases (b) Walking - TakingPhoto (c) Smoking - Phoning (d) Smoking - Sitting (e) Discussion - Directions (f) w/o Sitting - Sitting}
    \label{fig:roc-all}
\end{figure*}

Evaluating the quality of epistemic uncertainty is difficult due to the unavailability of ground truth annotations.
In the main paper, we conducted an experiment on classifying walking-related and sitting-related actions. Here, we subsequently present the results of further assessments of a broader range of actions in \Cref{tab:auroc-all} and detailed ROC curves in \Cref{fig:roc-all}. Our experimental results indicate that our proposed method outperformed previous techniques in almost all test scenarios. Specifically, in the last column of \Cref{tab:auroc-all}, we trained our clustering and forecasting model on all non sitting-related actions (13 actions) and evaluated EpU's ability to classify these actions from the remaining sitting-related actions (2 actions). Our method achieved higher AUROC and better ROC curves.

\begin{table*}[!h]
    \centering
    \begin{tabular}{lcccccc}
        \toprule
        Train actions & Walking & Walking & Smoking & Smoking & Discussions & w/o Sitting\\
        Test actions & Purchases & TakingPhoto & Phoning & Sitting & Directions & Sitting \\
        \midrule
        Deep-Ensemble-3 & 0.83 & 0.80 & 0.51 & 0.69 & 0.54 & 0.68\\ 
        Deep-Ensemble-5 & 0.86 & 0.80 & 0.54 & \textbf{0.77} & 0.58 & 0.68\\ 
        MC-Dropout-5 & 0.80 & 0.80 & 0.48 & 0.65 & 0.51 & 0.54\\
        MC-Dropout-10 & 0.83 & 0.82 & 0.49 & 0.67 & 0.52 & 0.55\\
        Ours & \textbf{0.93}& \textbf{0.89} & \textbf{0.56} & 0.71 & \textbf{0.59} & \textbf{0.76}\\ 
        \bottomrule
    \end{tabular}
    \caption{AUROC for different sets of actions for different epistemic uncertainty quantification methods. The actions on top indicate the train actions and the ones below them indicate the test actions.}
    \label{tab:auroc-all}
\end{table*}

\subsubsection{Motion clustering}

To derive EpU, we opted to use clustering in the representation space instead of alternative methods, such as action recognition models or the action labels of existing human pose datasets, e.g., Human3.6M.
There are several differences between a motion and an action when dealing with human pose sequences: 1) the actions are limited, whereas motions can be more varied; 2) multiple consecutive distinct motions usually constitute an action.
Motion clustering is also generalizable to datasets without action labels and real-world settings.

\end{document}